\documentclass[conference]{IEEEtran}
\IEEEoverridecommandlockouts
\usepackage{amsmath,amssymb,amsfonts}
\usepackage{algorithmic}
\usepackage{graphicx}
\usepackage{textcomp}
\usepackage{xcolor}

\usepackage{url}
\usepackage{multirow}

\def\BibTeX{{\rm B\kern-.05em{\sc i\kern-.025em b}\kern-.08em
    T\kern-.1667em\lower.7ex\hbox{E}\kern-.125emX}}
\begin{document}

\title{SPEAK: Speech-Driven Pose and Emotion-Adjustable
Talking Head Generation\\
\thanks{\textsuperscript{*}The authors contributed equally.}
\thanks{\textsuperscript{$\dagger$}Corresponding author: zeurdfish@gmail.com}
}

\author{\IEEEauthorblockN{Changpeng Cai\textsuperscript{*} }
\IEEEauthorblockA{\textit{Ping An Technology} \\
\textit{Southeast University} \\
Nanjing, China}
\and
\IEEEauthorblockN{Guinan Guo\textsuperscript{*}}
\IEEEauthorblockA{\textit{Ping An Technology} \\
\textit{Sun Yat-sen University} \\
Guangzhou, China}
\and
\IEEEauthorblockN{Jiao Li\textsuperscript{*} }
\IEEEauthorblockA{\textit{Ping An Technology} \\
Nanjing, China}
\and
\IEEEauthorblockN{Junhao Su\textsuperscript{*} }
\IEEEauthorblockA{\textit{Southeast University}\\
Nanjing, China}
\and
\IEEEauthorblockN{Fei Shen}
\IEEEauthorblockA{\textit{Nanjing University of} \\ 
\textit{Science and Technology} \\
Nanjing, China}

\and
\IEEEauthorblockN{Chenghao He}
\IEEEauthorblockA{\textit{East China University of} \\ 
\textit{Science and Technology} \\
Shanghai, China}
\and
\IEEEauthorblockN{Jing Xiao}
\IEEEauthorblockA{\textit{Ping An Technology}\\
Shenzhen, China}
\and
\IEEEauthorblockN{Yuanxu Chen}
\IEEEauthorblockA{\textit{Ping An Technology}\\
Shenzhen, China}
\and
\IEEEauthorblockN{Lei Dai}
\IEEEauthorblockA{\textit{Ping An Technology}\\
Nanjing, China}
\and
\IEEEauthorblockN{Feiyu Zhu\textsuperscript{$\dagger$}}
\IEEEauthorblockA{\textit{University of Shanghai for}\\ \textit{Science and Technology}\\
Shanghai, China}

}

\maketitle

\begin{abstract}
Most earlier researches on talking face generation have focused on the synchronization of lip motion and speech content. However, head pose and facial emotions are equally important characteristics of natural faces. While audio-driven talking face generation has seen notable advancements, existing methods either overlook facial emotions or are limited to specific individuals and cannot be applied to arbitrary subjects. In this paper, we propose a novel one-shot \textit{Talking Head} Generation framework (SPEAK) that distinguishes itself from the general \textit{Talking Face} Generation by enabling emotional and postural control. Specifically, we introduce Inter-Reconstructed Feature Disentanglement (IRFD) module to decouple facial features into three latent spaces. Then we design a face editing module that modifies speech content and facial latent codes into a single latent space. Subsequently, we present a novel generator that employs modified latent codes derived from the editing module to regulate emotional expression, head poses, and speech content in synthesizing facial animations. Extensive trials demonstrate that our method ensures lip synchronization with the audio while enabling decoupled control of facial features, it can generate realistic talking head with coordinated lip motions, authentic facial emotions, and smooth head movements. The demo video is available: \url{https://anonymous.4open.science/r/SPEAK-8A22}.
\end{abstract}


\section{Introduction}
Talking face generation is a critical technology requirement for multimedia applications~\cite{edwards2016jali}. One of the most challenging aspects of this undertaking is that human speech is frequently accompanied by non-linguistic elements, such as head posture and facial emotions ~\cite{williams1972emotions}, which are important for talking face generation. Researchers have been using analogous methods in this approach for many years because of the rapid development of deep learning and the widespread use of related technology. Modern technologies can simulate lip movements that are perfectly synchronized with the audio speech~\cite{chen2019hierarchical,prajwal2020lip}; however, the faces in such videos are generally emotionless. Most prior studies~\cite{chung2017you,suwajanakorn2017synthesizing,zhou2019talking,song2022everybody,ma2023dreamtalk}, have chosen to maintain the original pose in a video. Thus how to make talking face in generated video to expressive emotion is still an open problem. 
  
  In real-world scenarios, when different individuals speak the same words, their facial emotions and head poses come with individual stylistic features. Even for the same person, the facial emotions and head poses can vary when speaking the same sentence in different situations. Due to these pronounced diversities, the creation of talking heads with controllable head poses and emotions remains a significant challenge. In previous works Chen et al.~\cite{zhou2021pose,prajwal2020lip} can only control one of facial emotions and head motion. Some recent methods~\cite{ma2023styletalk,ji2022eamm} adopt multiple sources of information as intermediate representations, but the quality of pose control and facial generation is limited.~\cite{zhua2023audio} shows that diffusion models are stable, but the generated faces lack emotion. Thus, controlling the speaker's head pose and facial emotions end-to-end is highly desirable.
  
  To address these challenges, this paper presents a pioneering framework distinct from general Talking {\bfseries Face} Generation, known as \underline{S}peech-Driven \underline{P}ose and \underline{E}motion-\underline{A}djustable Tal\underline{k}ing {\bfseries Head} Generation (SPEAK). Our objective is to create realistic talking videos utilizing four input types: an identity source image exhibiting a neutral expression, a spoken source audio, a pose source video, and an emotion source video.

  In detail, we initially disentangle the identity image, pose video, and emotion video into their corresponding latent spaces through an innovative Inter-Reconstructed Feature Disentanglement (IRFD) module. To amalgamate facial identity features, head pose features, mouth movements, and local emotion representations. we empirically explored the intrinsic mechanisms of facial motion. We identified a compelling trait: the emotion representations encompass a portion of the mouth movement representations. Consequently, we designed an editing module to align and merge these features. Ultimately, an image generator employs the aligned features as input to produce the talking head.
  
  Our contributions are summarized as follows: 
  
  \textbf{1)}We propose Speech-Driven Pose and Emotion-Adjustable Talking Head Generation (SPEAK) framework, which is an innovative attempt for GAN-based talking head generation methods to achieve better pose and emotion control. 
  
  \textbf{2)}We present a self-supervised Inter-Reconstructed Feature disentanglement(IRFD) module that can independently distill emotion, identity and pose information to latent space from reference single frame. We do not need to do substantial preprocessing on the training data because the emotion and head poses are learned in the implicit space. 
  
  \textbf{3)}We introduce an editing module, to align speech content latent codes with facial emotions embedding and combine it with identity and posture information embedding, resulting in more accurate and realistic audio-driven talking head.

\section{Method}
\begin{figure*}[t]
  \centering
  \includegraphics[scale=0.28]{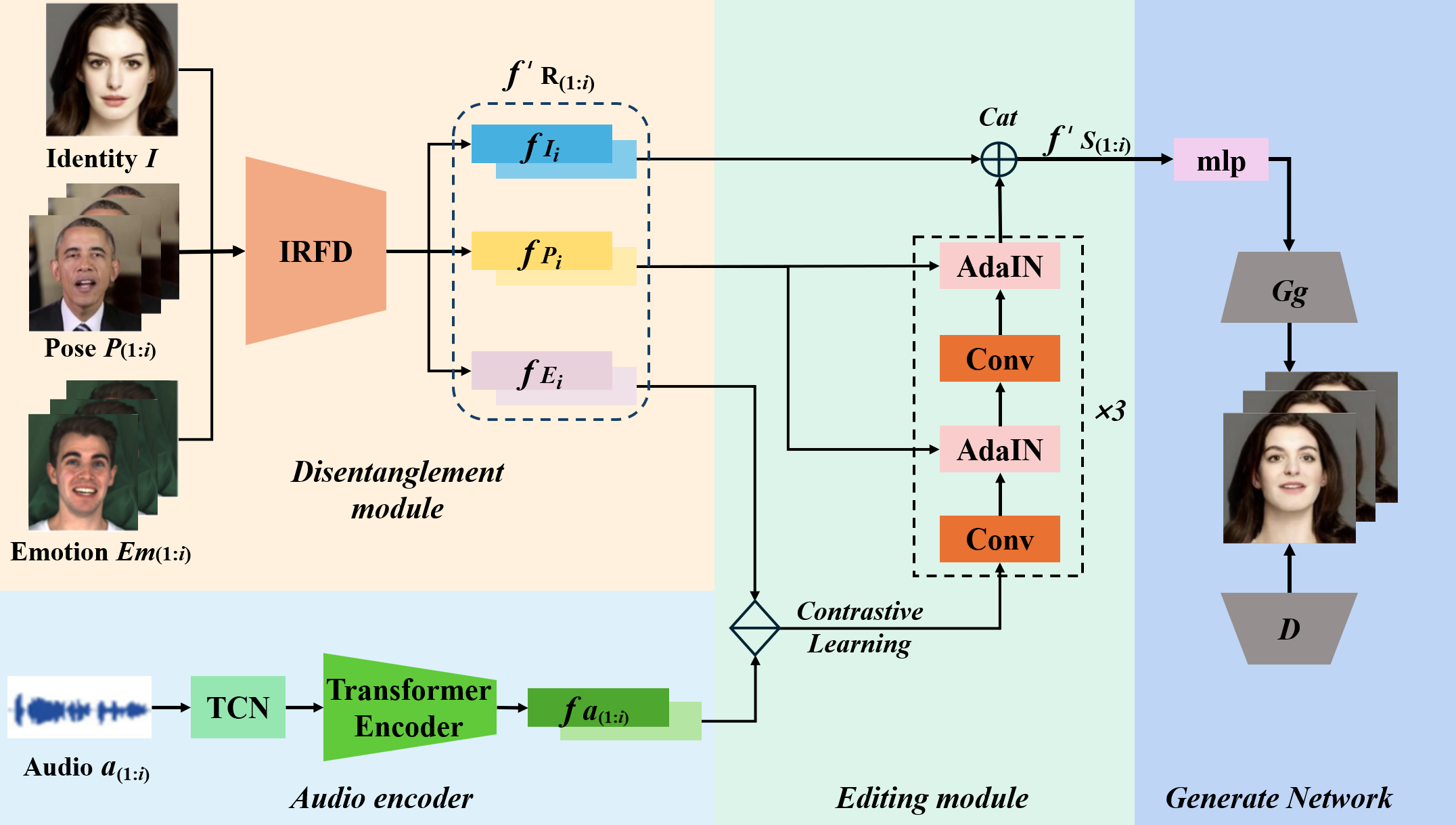}
  \caption{Illustration of our proposed Talking Head Generation Framework. Our framework first extracts human face. To begin with, We employ the IRFD to decouple facial features from video clips $I$,$P_{(1:i)}$,$Em_{(1:i)}$ onto three latent spaces$f_{I_{i}}$, $f_{P_{i}}$, $f_{E_{i}}$. An audio encoder encodes speech wavform into audio content features $f_{a_{(1:i)}}$. Then the editing module aligning audio content $f_{a_{(1:i)}}$ and facial information $f_{R_{(1:i)}}^{'}$ modalities.}
  \label{overflow}
\end{figure*}

\subsection{Talking Head Generation}
\noindent \textbf{Disentanglement Module.} To generate a talking face with controlled pose and facial emotion, the expression, pose, and identity features must be decoupled separately from the raw spaces of the latent face.  We adopt three independent encoders in IRFD module to decompose high-level facial features into three low-level latent feature spaces, which reflect head pose, facial emotion, and identity.

\noindent \textbf{Audio Encoder.} Our method utilizes the state-of-the-art (SOTA) self-supervised pre-trained speech model as the architecture for our speech encoder. The model is comprised of an audio feature extractor and a multi-layer transformer encoder. The audio feature extractor leverages a temporal convolutional network (TCN) to transform the speech raw waveform into feature vectors. The transformer encoder then generates contextualized speech representations from the audio features by utilizing an effective attention scheme.

\noindent \textbf{Editing Module.} To compensate the information loss and fusion audio features with image features, we design an editing module. As shown in Fig. \ref{overflow}, the global audio vectors $f_{a_{(1:i)}}$ are sent into editing module along with the disentangled emotion embedding $f_{E_{i}}$. Then at different levels of the network, we introduce random noise and inject the facial features codes $f^{'}_{R_{(1:i)}}$ by AdaIN blocks~\cite{suvorov2022resolution} which normalize visual features channel-wise after each fully connected block. So the multimodal latent output $f^{'}_{S_{(1:i)}}$ from editing module can help capture subtle image details at different resolutions~\cite{karras2020analyzing} and further generate realistic speech-driven talking head with diverse styles. 

\noindent \textbf{Generate Network.} The facial information of emotion and pose, and speaking content of audio clip have already been edited in the latent space. Since IRFD capture the head pose and emotion without global audio features while audio-driven talking head generation, we use two individual generators for better interpretation of the edited latent codes, i.e., the IRFD generator and global generator $G_g$. For our generating network, we novelty modify the head generator based on styleGAN~\cite{karras2020analyzing} for the two different generation scenarios. As shown in Fig. \ref{overflow}, we feed the edited talking head latent codes into the generator to generate a talking head with synchronized lips, emotions, and poses. Specially, at convolutional block, we add multi-layer perceptron (mlp) results of $f^{'}_{R_{(1:i)}}$ to map the facial information. So the global generator can be defined as:
 \begin{equation}
 \mathcal I_{g_i}=G_{g}(mlp(f_{I_{i}},f_{P_{i}}),\mathcal F),
 \end{equation}
 where $\mathcal F=f^{'}_{S_{(1:i)}}$ and $\mathcal I_{g_i}$ is the output image of the Audio-driven Talking Head Generator. While pre-training the disentanglement module, the IRFD generator is $\mathcal I_{d_i}=G_{d}(mlp(f_{I_{i}},f_{P_{i}},f_{E_{i}}))$.

 \subsection{Network Training}
The following is the training procedure. We choose a identity frame $I$ in the beginning and extract its identity embedding. The input video $R_{(1:M)}$ is used to extract the emotion and pose embedding. $E_{a}$ converts the appropriate audio clip to an audio waveform and extracts its speech embedding. The speech embedding is then combined with image embeddings (identity, emotion, pose) by editing module. Finally, this edited latent codes are fed into generator $G_g$. The reconstructed video is $\mathcal I_{d_(1:M)}$. A multiscale discriminator $D$ is applied to the generated video frame and original video frame to evaluate whether the video frames are fake or real. 
Therefore, we constrain this by introducing the single-image adversarial loss $\mathcal L_{GAN}$ reported in~\cite{song2022everybody}:
\begin{equation}
\begin{aligned}
\mathcal L_{GAN} & =\mathop{min}\limits_{G}\mathop{max}\limits_{D}E_{(x,y)}[logD(x,y)]\\&+E_{x}[log(1-D(x,G(x)))],
\end{aligned}
\end{equation}

\noindent where $x$ is the input identity frame$I$ and $y$ is the original video frame of video $R_{(1:M)}$.

To achieve a more realistic talking head, we use contrastive loss~\cite{chung2017out} to enhance the synchronization of audio and visual elements. We propose, based on recent research~\cite{eskimez2020end}, to adopt and train a modified version of SyncNet~\cite{chung2017out}:
\begin{equation}
\begin{aligned}
\mathcal L_{sync}&=\frac{1}{2N}\sum_{n=1}^{N} (y_{n})d_{n}^{2}\\& +(1-y_{n})max(margin-d_{n},0)^{2},y\in[0,1],
\end{aligned}
\end{equation}

\noindent where $y$ is the binary similarity metric between the audio $a_{(1:M)}$ and the video inputs $R_{(1:M)}$. $d$ is the $L2$ distance between the audio and video embedding:
\begin{equation}
d=||Syncnet_{fc7}(R_i)-Syncnet_{fc7}(\mathcal I_{g_i})||_{2},
\end{equation}



Then, by combining the perceptual reconstruction loss $\mathcal L_{vgg}$~\cite{isola2017image}, which is calculated by comparing pre-trained VGGNet\cite{kollias2019deep} features from different layers of the network, the final loss function for the generator is:
\begin{equation}
\mathcal L=\mathcal L_{GAN}+\mathcal L_{sync}+\mathcal L_{vgg}.
\end{equation}

\begin{figure}[ht]
  \centering
  \includegraphics[scale=0.63]{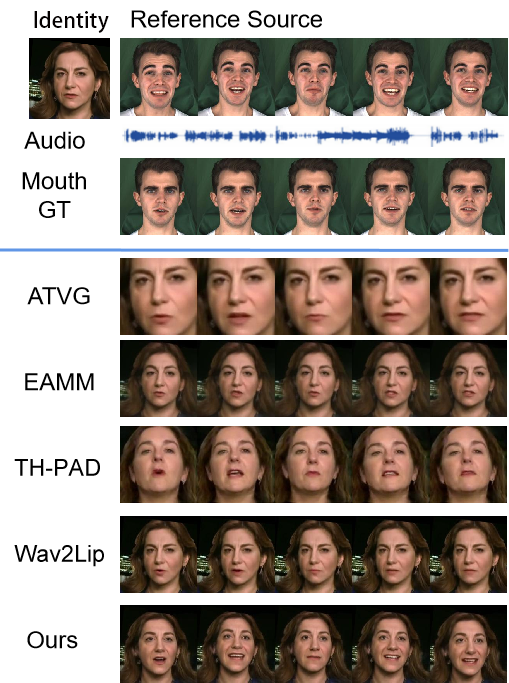}
  \caption{Qualitative comparisons with other baselines. The top two rows show the Identity, Reference Source (video frames after the fusion of emotion and pose) and Audio. Since it is the first method to generate videos using four types of input data, there's no prior ground-truth with all four inputs. We combine Audio and Reference Source to generate mouth shape as Mouth GT.}
  \label{tf2}
\end{figure}

\begin{table*}
\caption{Quantitative results on MEAD ~\cite{wang2020mead} and HDTF ~\cite{zhang2021flow}. \textbf{Bold} means the best.}
\centering
\label{table1}
\resizebox{\textwidth}{!}{%
\begin{tabular}{l|lllll|lllll} 
\hline
\multirow{2}{*}{Method / Score} & \multicolumn{5}{c|}{MEAD}           & \multicolumn{5}{c}{HDTF}             \\
                        & SSIM$\uparrow$ & PSNR$\uparrow$  & F-LMD$\downarrow$ & M-LMD$\downarrow$ & Sync$_{conf}$$\uparrow$ & SSIM$\uparrow$ & PSNR$\uparrow$  & F-LMD$\downarrow$ & M-LMD$\downarrow$ & Sync$_{conf}$$\uparrow$  \\ 
\hline
ATVG \cite{chen2019hierarchical}          & 0.73 & 29.33 & 3.36  & 5.82  & 3.81 & 0.67 & 28.13 & 3.22  & 4.16  & 3.93  \\
EAMM \cite{ji2022eamm}      & 0.44 & 28.54 & 5.78  & 5.83  & 1.54 & 0.36 & 27.66 & 6.04  & 5.92  & 1.72  \\
TH-PAD \cite{yu2023talking}               & 0.79 & 29.07 & \textbf{2.26 } & 3.17  & 3.53 & 0.78 & 29.19 & 2.35  & 2.74  & 4.88  \\
Wav2Lip \cite{prajwal2020lip}               & 0.69 & 29.01 & 3.57  & 3.36  & \textbf{5.13} & 0.48 & 27.98 & 4.77  & 4.24  & 5.73  \\
\hline
Ours                    & \textbf{0.84} & \textbf{29.41} & 2.48  & \bfseries{2.92}  & 4.99 & \bfseries{0.83} & \bfseries{29.46} & \textbf{2.32}  & \textbf{2.42}  & \textbf{5.76}  \\
Ground Truth             & 1    & -     & 0     & 0     & 5.26 & 1    & -     & 0     & 0     & 5.83  \\ 
\hline
\end{tabular}
}
\label{Quantitative}
\end{table*}

\section{Experiment}

\noindent \textbf{Datasets.} We use Voxceleb~\cite{nagrani2017voxceleb}, MEAD~\cite{wang2020mead}, CREMA-D~\cite{cao2014crema} and HDTF~\cite{zhang2021flow} datasets on our experiments. Voxceleb contains real world videos with large variation in face pose and occlusion of faces and audio. MEAD is a high-quality emotional audiovisual dataset that includes 60 actors/actresses and eight emotion categories. CREMA-D is a dataset with over 7,000 emotion category audios. HDTF is a high-resolution in-the-wild audio-visual dataset. We crop faces from the videos and resize them to $256 \times 256$. We train our IRFD on the Voxceleb and MEAD datasets and train other SPEAK modules on the Voxceleb and MEAD datasets.

\noindent \textbf{Implementation Details.} We employ Adam optimizer for training. The structure of encoders in IRFD is ResNet50 \cite{he2016deep}. Before training the SPEAK, the encoders and generator of IRFD are trained to encode facial features of pose, emotion, and identity into latent spaces. We randomly swap one type of facial feature code and fed into the IRFD generator to reconstruct full face. Then IRFD generator is discarded, while its encoders are used in the global generator $G_g$ training process. All models are trained on 8 NVIDIA V100 GPUs.

\noindent \textbf{Evaluation Metrics and Baselines.} We use peak signal-to-noise ratio (PSNR) and structural similarity (SSIM)~\cite{wang2004image} to compare the image quality of the generated video frames to the original video frames. To account for the accuracy of mouth forms and lip sync, we employ both the landmark distance (LMD) surrounding the mouths (M-LMD)~\cite{chen2019hierarchical} and the confidence score (Sync$_{conf}$)~\cite{chung2017out} to measure audiovisual synchronization. Then we use the LDM on the whole face (F-LMD) to measure the accuracy of facial emotions and pose.We compared our SPEAK to ATVG \cite{chen2019hierarchical}, EAMM \cite{ji2022eamm}, SOTA GAN-based method TH-PAD \cite{yu2023talking} and Wav2Lip \cite{prajwal2020lip}.

\noindent \textbf{Quantitative Evaluation.} As results shown in Table \ref{Quantitative}, our method outperforms most metrics across both datasets. Since Wav2Lip \cite{prajwal2020lip} merely generated lip-sync and does not change other parts of the reference images, it obtains the highest Sync$_{conf}$ on MEAD. And our Sync$_{conf}$ is closest to ground truth on MEAD and the highest on HDTF dataset. Besides, our method significantly outperforms the SOTA GAN-based method TH-PAD \cite{yu2023talking}. These indicates the superior quality of our generated real face (measured by PSNR and SSIM) as well as a highly accurate lip synthesis (measured by M-LMD and Sync$_{conf}$) and emotions and pose (measured by F-LMD).

\noindent \textbf{Qualitative Evaluation.} Result in Fig. \ref{tf2} is evident that our SPEAK can achieve high-fidelity emotions and posture styles matching the Reference Source, as well as more accurate lip shapes. In terms of lip-sync, Wav2Lip \cite{prajwal2020lip} and ATVG \cite{chen2019hierarchical} don't consider emotions, so they generate seemingly reasonable lip movements but only with neutral emotions. EAMM \cite{ji2022eamm} and TH-PAD \cite{yu2023talking} cannot achieve accurate lip-sync and pose. In contrast, SPEAK can imitate emotional talking head from reference source driving video clips while achieving a accurate lip-sync, satisfactory identity preservation. 

\begin{table}
\caption{User study on CREMA-D~\cite{cao2014crema} are measured using mean scores.}
\centering
\label{table2}
\resizebox{\linewidth}{!}{%
\begin{tabular}{l|ccc} 
\hline \multirow{2}{*}{Method / Score}& \multicolumn{3}{c}{CREMA-D}\\
& Degree of Lip-Sync $\uparrow$ & Head Naturalness $\uparrow$ & Video Realness $\uparrow$ \\
\hline 
ATVG \cite{chen2019hierarchical} & 4.21 & 3.51 & 2.58 \\
EAMM \cite{ji2022eamm} & 3.97 & 1.56 & 1.86 \\
TH-PAD \cite{yu2023talking} &\underline{3.71} & 3.56 & 3.85 \\
Wav2Lip \cite{prajwal2020lip}  & 3.82 & \underline{1.31} & \underline{1.58} \\
\hline 
Ours & \textbf{4.23} & \textbf{3.67} &  \textbf{4.11} \\
Ground Truth & 4.88 & 4.89 & 4.96 \\
\hline
\end{tabular}
}
\label{tableusers}
\end{table}

\noindent \textbf{User Study.} We conduct an user study with 20 participants to gather human feedback on 30 videos generated by SPEAK and other methods. After shuffling generated videos, users need to score each video on a scale of 1 (bad) to 5 (good) based on (1) degree of lip-sync, (2)naturalness of head motions, and (3) video realness. Users give the highest marks on each aspect of our method in Table \ref{tableusers}. TH-PAD \cite{yu2023talking} is highly competitive in terms of video realness and naturalness of head motions, but its lip-sync performance is worst. The head naturalness and video realness of both Wav2Lip \cite{prajwal2020lip} and EAMM \cite{ji2022eamm} are relatively poor, while the video realness of ATVG \cite{chen2019hierarchical} is not perfect compare to our proposed method. 

\begin{figure}
  \centering
  \includegraphics[width=\linewidth]{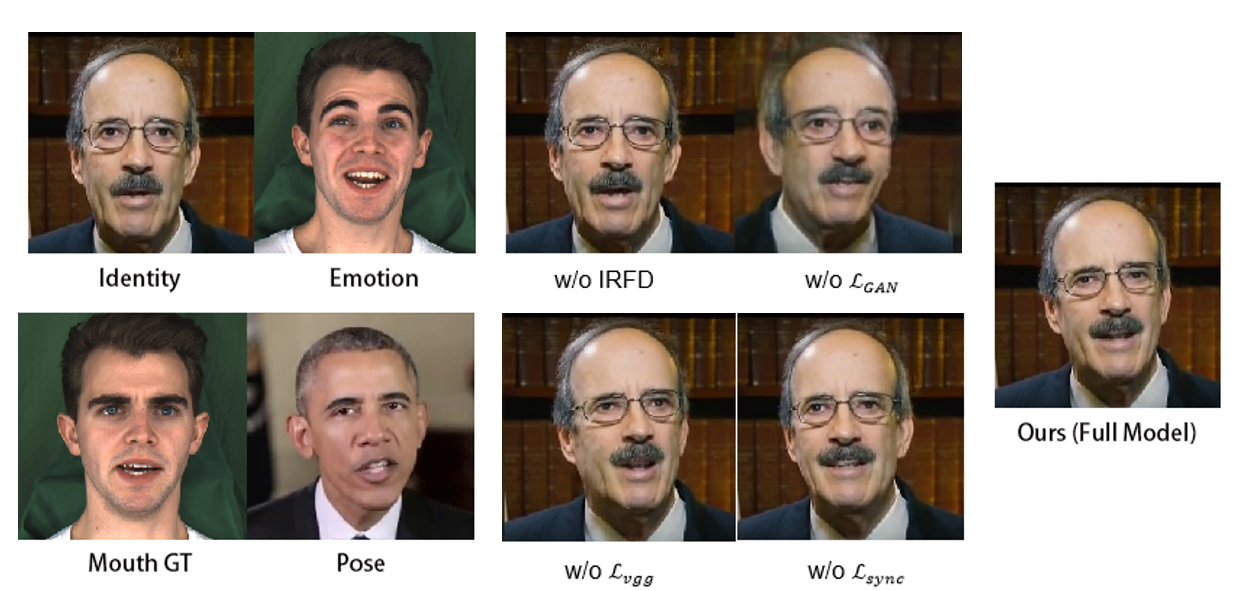}
  \caption{Visualized results of SPEAK ablation study.}
  \label{ablation}
\end{figure}

\noindent \textbf{Ablation study.}
We provide a detailed analysis of the impact of each module on the MEAD dataset. Quantitative results for each module of IRFD in Table \ref{ablation_SPEAK} indicate that the IRFD Identity encoder contributes the most to SSIM and PSNR, the IRFD Pose encoder contributes most to F-LMD, and the IRFD Emotion encoder contributes most to M-LMD and Sync$_{conf}$. This implies that IRFD module achieves our design motivation. The visualized results in Fig. \ref{ablation} show the impact of each module in the entire SPEAK. When $\mathcal{L}_{sync}$ is removed, it is difficult to generate faces synchronized with mouth GT. Although the performance after removing $\mathcal L_{vgg}$ or $\mathcal L_{GAN}$ is also quite competitive, our method imitates the eye angle in the pose source more accurately. When IRFD module is removed, the less accurate qualitative findings are hard to achieve better pose and emotion control.

\begin{table}
\caption{Ablation study on quantitative comparison on MEAD~\cite{wang2020mead}.}
\centering
\resizebox{\linewidth}{!}{
\begin{tabular}{l|lllll}
\hline
\multirow{2}{*}{Method / Score} & \multicolumn{5}{c}{MEAD} \\
                        & SSIM$\uparrow$ & PSNR$\uparrow$  & F-LMD$\downarrow$ & M-LMD$\downarrow$ & Sync$_{conf}$$\uparrow$ \\ 
\hline
w/o $\text{IRFD}_{\text{Identity}}$                    & \underline{0.53}	&\underline{23.89}	&4.22	&3.84	&4.11\\
w/o $\text{IRFD}_{\text{Pose}}$                    &0.57	&24.77	&\underline{5.15}	&3.78	&3.98\\
w/o $\text{IRFD}_{\text{Emotion}}$                    &0.62	&25.31	&3.17	&\underline{4.72}	&\underline{3.23}\\



\hline
Ours (Full Model)                    
& \textbf{0.84} & \textbf{29.41} & \textbf{2.48}  & \textbf{2.92}  & \textbf{4.99}\\
\hline
\end{tabular}
}
\label{ablation_SPEAK}
\end{table}

\section{Conclusion}
 In this paper, we propose a technique to generate accurately lip-synched emotional talking head with free pose and emotion control from other videos. We design a novel disentanglement module IRFD for decomposing input sample into emotion, identity, and pose embedding. Then we provide a novel talking head generation framework SPEAK. Qualitative and quantitative experiments indicate that our method performs very robustly in challenging scenarios, such as significant pose and emotional expression variations.

\bibliographystyle{IEEEtran}
\bibliography{conference_101719}

\end{document}